\documentclass[11pt]{article}

\usepackage[final]{acl}

\usepackage{times}
\usepackage{latexsym}

\usepackage[T1]{fontenc}
\usepackage[utf8]{inputenc}

\usepackage{microtype}

\usepackage{inconsolata}

\usepackage{graphicx}

\usepackage{amsmath}
\usepackage[normalem]{ulem} % 提供 \dotuline 命令
\usepackage{multirow}       % 用于多行表头
\usepackage{booktabs}       % 用于表格线条
\usepackage{amssymb}
\usepackage{xcolor}

\usepackage{CJKutf8}

\usepackage{enumitem}

\title{Beyond Cosine Similarity: Zero-Initialized Residual Complex Projection for Aspect-Based Sentiment Analysis}

\author{
  Yijin Wang\textsuperscript{1} \and
  Fandi Sun\textsuperscript{1} \and
  Haoyu Wen\textsuperscript{1}\thanks{Corresponding author: hywen@xidian.edu.cn} \\
  \textsuperscript{1}School of Economics and Management, Xidian University, Xi'an, China \\
  \texttt{hywen@xidian.edu.cn}
}

\begin{document}
\maketitle

% \setlength{\abovedisplayskip}{0pt}  % 公式上方的间距
% \setlength{\belowdisplayskip}{0pt}  % 公式下方的间距
% % 以下两项通常不需要调整
% \setlength{\abovedisplayshortskip}{0pt}
% \setlength{\belowdisplayshortskip}{0pt}

% ==========================================
% abstract
% ==========================================

\begin{abstract}
Aspect-Based Sentiment Analysis (ABSA) faces critical challenges due to representation entanglement and false-negative collisions in real-valued embedding spaces. In this paper, we propose a novel framework featuring a Zero-Initialized Residual Complex Projection (ZRCP) and an Anti-collision Masked Angle Loss. Our approach projects textual features into a complex semantic space, utilizing the phase to isolate sentiment polarities while regularizing the amplitude to ensure structural consistency within aspect categories. To mitigate this, we introduce an anti-collision mask that preserves intra-polarity aspect cohesion while \textbf{significantly expanding the discriminative margin between opposing polarities}. Experimental results on the ASAP dataset demonstrate that our framework achieves a state-of-the-art Macro-F1 score of 0.8923, outperforming robust baselines. 
\end{abstract}

% ==========================================
% 1. Introduction
% ==========================================

\begin{figure*}[htbp]
    \centering
    \includegraphics[width=0.9\textwidth]{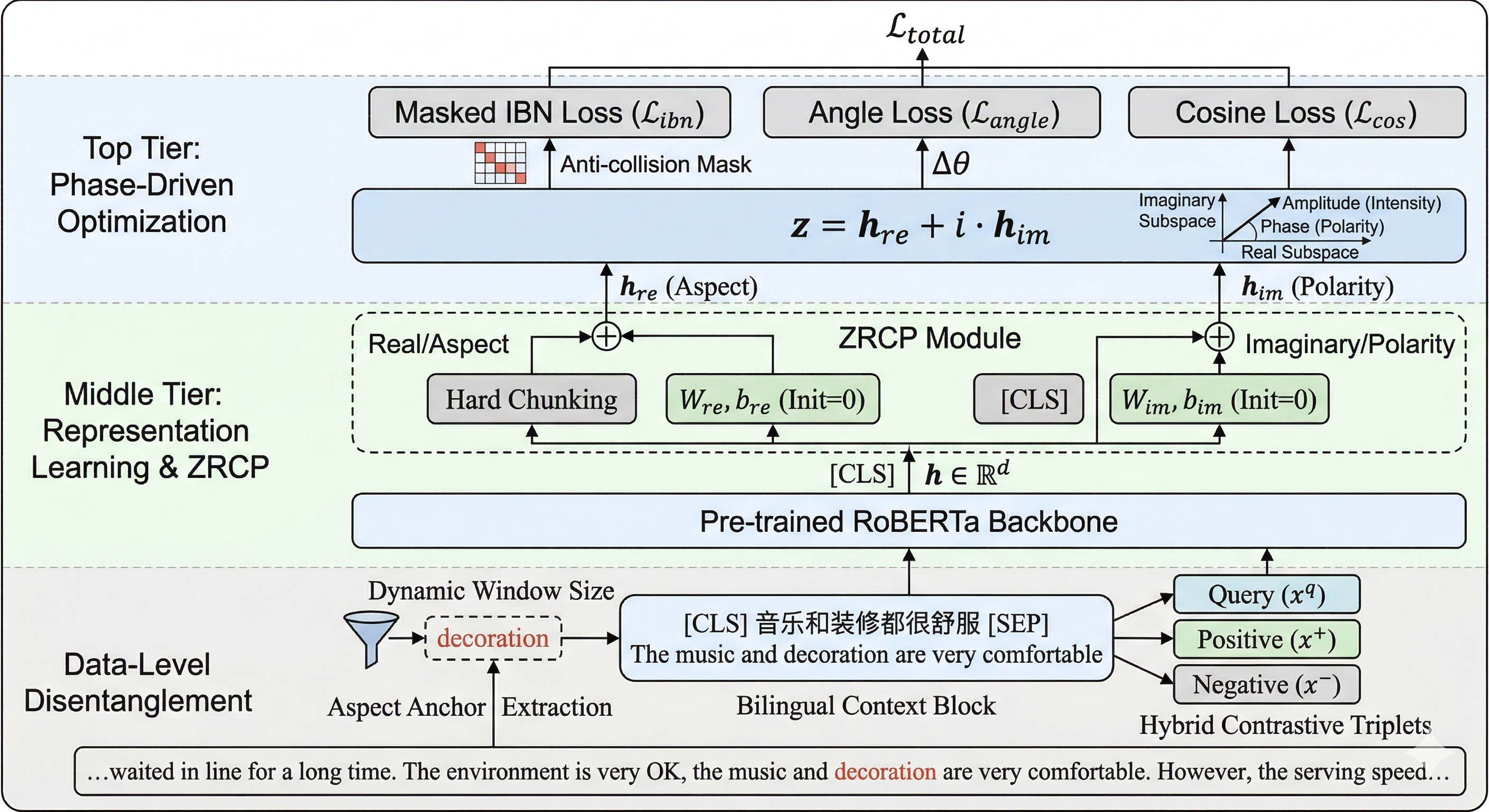}
    \caption{Overall architecture of our Phase-Driven Disentanglement framework. Textual inputs are mapped into a complex space via the ZRCP module and optimized using an Anti-collision Masked Angle Loss to decouple aspect and sentiment.}
    \label{fig:main_architecture}
\end{figure*}

\section{Introduction}
Aspect-Based Sentiment Analysis (ABSA) aims to extract specific aspect categories and sentiment polarities from unstructured reviews. Mapping multi-aspect sentences into standard real-valued spaces frequently causes representation entanglement, conflating objective aspect semantics and subjective sentiment polarities\cite{nazir2020issues}\cite{schouten2015survey}. 

Recent advancements have leveraged contrastive learning, such as InfoNCE, to carve out better discriminative margins in the feature space\cite{xu2024contrastive}. However, we observe that applying standard contrastive learning directly to fine-grained ABSA tasks paradoxically induces severe \textit{false-negative collisions}. Standard instance discrimination erroneously repels sentences sharing identical aspects and polarities (e.g., two positive reviews about "Taste"), destroying intra-aspect cohesion and blurring decision boundaries.

To overcome this representation constraint, we argue that complex-valued geometry offers a natural fit for ABSA. Unlike real-valued embeddings that conflate attributes into a single scalar dimension, complex numbers inherently decouple directional and magnitude properties. In our formulation, the \textit{phase} (angle) isolates subjective sentiment polarities—which are inherently oppositional—while the \textit{amplitude} (magnitude) captures semantic intensity or lexical richness\cite{zhao2024qpen}. Crucially, this separation allows for explicit regularization of the amplitude to filter out intensity-related noise that often interferes with polarity judgment.

Based on this geometric intuition, we propose a novel framework featuring the \textbf{Zero-Initialized Residual Complex Projection (ZRCP)}. The ZRCP module smoothly maps pre-trained real-valued features into a complex semantic space\cite{wang2019encoding}. By anchoring the resulting amplitudes through a dedicated penalty, we prevent lexical variance from blurring the discriminative boundaries of sentiment phases.

Building upon this complex representation, we design an \textbf{Anti-collision Masked Angle Loss}. By introducing a dynamic mask matrix, our loss function prevents the erroneous repulsion of same-polarity instances within the same aspect cluster. This enables the model to maximize the inter-polarity angular margin without sacrificing global aspect cohesion.

The main contributions of this paper are summarized as follows:
\begin{itemize}[nosep, leftmargin=*]
    \item We propose a novel ZRCP module that projects textual embeddings into a complex space, providing a mathematical foundation for disentangling objective aspects and subjective polarities. 
    \item We introduce an Anti-collision Masked Angle Loss that effectively addresses this erroneous repulsion, expanding the relative inter-polarity discriminative margin by over 50\% without destroying structural cohesion.
    \item We investigate the geometric role of complex amplitude and propose an \textbf{Amplitude Penalty ($\mathcal{L}_{amp}$)} that achieves structural consolidation. We prove that this regularization effectively filters intensity-related noise, significantly boosting performance on class-imbalanced aspects. 
    \item Extensive experiments demonstrate that our framework achieves a Macro-F1 score of 0.8923 on the ASAP dataset, significantly outperforming robust baselines.
\end{itemize}

% ==========================================
% 2. Related Work
% ==========================================
\section{Related Work}
\label{sec:related_work}

\subsection{Aspect-Based Sentiment Analysis}
Aspect-Based Sentiment Analysis (ABSA) identifies sentiment polarities towards specific aspect terms or categories within a sentence \cite{zhang2022survey}. Currently, Pre-trained Language Models (PLMs) like BERT and RoBERTa dominate ABSA by leveraging massive amounts of pre-training data \cite{xu2019bert,sirisha2022aspect}. Despite their strong performance, these models inherently map textual inputs into a continuous, real-valued high-dimensional space. In such spaces, objective aspect semantics (e.g., "Taste", "Decoration") and subjective sentiment polarities (e.g., "Positive", "Negative") are often heavily conflated. This representation entanglement restricts the model's capacity to draw crisp decision boundaries for highly subjective expressions.

\subsection{Contrastive Learning in NLP}
Contrastive learning has emerged as a powerful technique for learning discriminative sentence representations. Methods like SimCSE \cite{gao2021simcse} and its supervised variants have demonstrated that optimizing the InfoNCE loss can effectively pull semantically similar sentences together while pushing apart dissimilar ones, creating a more uniform and aligned embedding space. However, applying standard contrastive learning to fine-grained ABSA introduces a critical limitation: erroneous in-batch repulsion of same-class samples. In aspect-level datasets, sentences frequently share the exact same aspect category and sentiment polarity. Standard contrastive objectives, which are primarily designed for instance discrimination, erroneously treat these same-class samples in a batch as negative pairs, forcing them apart \cite{wei2025consistent}. This aggressive repulsion destroys intra-aspect cohesion and causes spatial collapse. Unlike previous works that attempt to re-weight negative samples, our framework introduces a dynamic \textit{Anti-collision Mask} that structurally preserves essential semantic cohesion while expanding the inter-polarity margin.

\subsection{Complex-Valued Representations}
Complex-valued neural networks have recently gained traction in NLP due to their superior capacity to model uncertainties and rich feature interactions \cite{lee2022complex,zhang2023quantum}. Complex numbers inherently possess two orthogonal degrees of freedom: the real part and the imaginary part. Prior works explored complex-valued embeddings to capture polysemy \cite{wang2019encoding}, but largely treated the complex domain as a black-box augmentation. Notably, AnglE \cite{li2024angleoptimizedtextembeddings} introduced a framework optimizing angular margins for general text embeddings. However, its direct hard-chunking strategy bisects the hidden state, which can disrupt pre-trained semantic continuity when applied to fine-grained tasks.

The distinct geometric roles of phase and amplitude remain underexplored for sentiment disentanglement \cite{zhao2022quantum}. In this paper, we propose the ZRCP module to explicitly exploit this phase-amplitude dichotomy. By systematically dedicating the phase to discrete sentiment polarities and the amplitude to continuous subjective intensity, our framework substantially alleviates the entanglement issue inherent in real-valued representations.

% ==========================================
% 3. Task Formulation & Data Construction
% ==========================================

\section{Task Formulation \& Data Construction}

\subsection{Context-Aware Aspect Extraction}
Lengthy reviews often contain multiple aspects with conflicting polarities (e.g., praising food while complaining about parking), exacerbating feature entanglement. To prevent the pre-trained language model from conflating these distinct sentiments, we perform a data-level disentanglement prior to representation learning\cite{wang2024disentangled}.

Specifically, we utilize a combination of TF-IDF scored keywords and an extended expert lexicon to precisely locate aspect-specific anchors within the raw reviews. To isolate the pure sentiment associated with a specific aspect, we propose a \textbf{Dynamic Window Size} algorithm\cite{rafiuddin2024exploiting}. Instead of feeding the entire review to the model, we extract a local context block surrounding the anchor word. The window size is dynamically adjusted based on the length of the targeted segment (ranging from 2 to 4 surrounding sentences) to ensure high-recall semantic completeness while strictly excluding extraneous aspect sentiments.

\subsection{Hybrid Aspect-Polarity Pair Generation}
To construct robust training signals for contrastive learning, we map the extracted context blocks into aspect-polarity sentence pairs. For a given anchor query ($x^q$), we pair it with a target sample ($x^t$) that shares the exact same objective aspect (e.g., ``Transportation''). We assign a label $y=1$ if $x^t$ shares the same polarity as the query (Positive Pair), and $y=0$ if it exhibits opposing polarity (Negative Pair). 

To enrich the diversity of the contrastive gradient signals, we implement a hybrid pair generation strategy:
\begin{itemize}[nosep, leftmargin=*]
    \item \textbf{Standard Pairs:} Mining target samples based on standard positive/negative polarity matching.
    \item \textbf{Data Augmentation:} Fusing multiple text blocks of the same polarity into a single augmented target to increase the density of the semantic space.
    \item \textbf{Hard Contrastive Mining:} Selecting positive pairs that share the same sentiment but exhibit high lexical variance, coercing the model to capture deep semantic alignment rather than superficial word overlap\cite{liang2021enhancing}.
\end{itemize}

% ==========================================
% 4. Methodology
% ==========================================

\section{Methodology}
\subsection{Contextual Encoding Backbone}
We encode input texts using the pre-trained \texttt{chinese-roberta-wwm-ext} \cite{2019RoBERTa} model. The hidden state of the \texttt{[CLS]} token from the final transformer layer serves as the aggregated sentence embedding, denoted as $h \in \mathbb{R}^d$, where $d$ is the hidden dimension.

\begin{figure*}[htbp]
    \centering
    \includegraphics[width=1.0\linewidth]{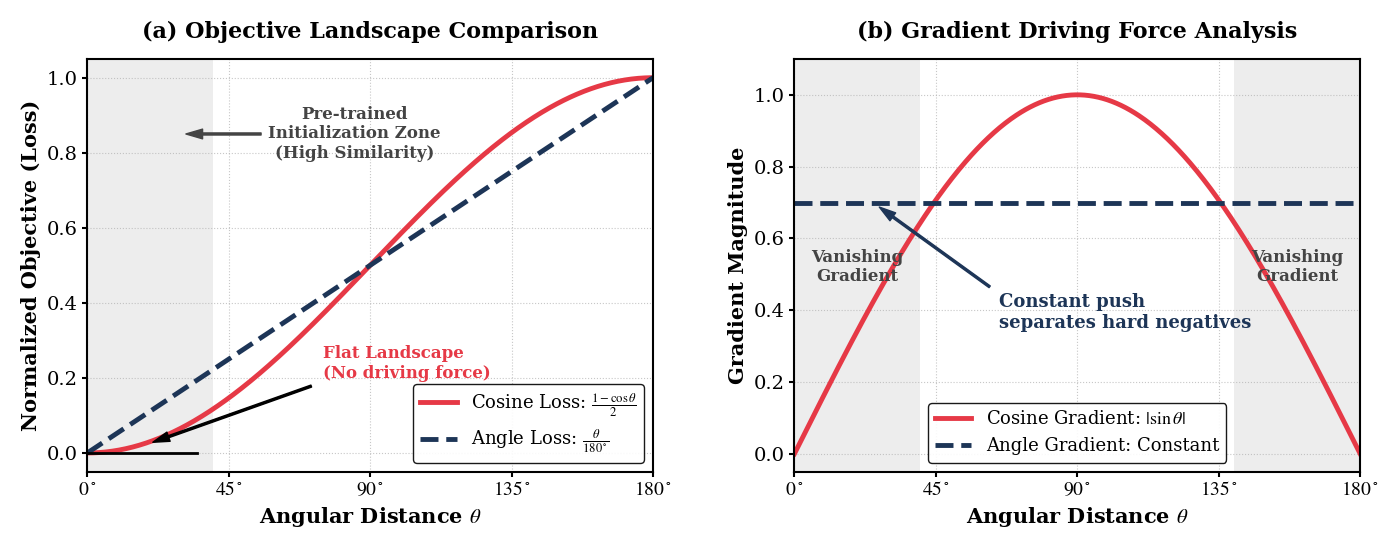}
    \caption{Theoretical comparison of Cosine and Angle Loss. \textbf{(a)} In the dense pre-trained initialization zone ($\theta \to 0^\circ$), the cosine objective saturates into a flat landscape. \textbf{(b)} This saturation causes severe vanishing gradients ($|\sin\theta| \to 0$). In contrast, our angle-based objective maintains a constant driving force to effectively separate entangled hard negatives.}
    \label{fig:gradient_vertical}
\end{figure*}

\subsection{Zero-Initialized Residual Complex Projection}
\label{sec:zrcp_module}

To map the real-valued backbone features $h \in \mathbb{R}^d$ into the complex semantic space $\mathbb{C}^{d/2}$, we propose the Zero-Initialized Residual Complex Projection (ZRCP). First, we bisect the feature $h$ into two equal-sized vectors $h^{(1)}, h^{(2)} \in \mathbb{R}^{d/2}$. To transform these into the real and imaginary parts without disrupting the pre-trained semantic topology, we apply a residual projection:
\begin{align}
h_{re} &= h^{(1)} + (W_{re}h^{(1)} + b_{re}) \label{eq:h_re} \\
h_{im} &= h^{(2)} + (W_{im}h^{(2)} + b_{im}) \label{eq:h_im}
\end{align}
where $W_{re}, W_{im} \in \mathbb{R}^{\frac{d}{2} \times \frac{d}{2}}$ are learnable weight matrices and $b_{re}, b_{im} \in \mathbb{R}^{\frac{d}{2}}$ are bias vectors. 

Crucially, we initialize all parameters in $\{W_{re}, b_{re}, W_{im}, b_{im}\}$ to \textbf{zero}. This zero-initialization strategy ensures that at the onset of training, $h_{re} \to h^{(1)}$ and $h_{im} \to h^{(2)}$. Consequently, the complex projection initially degenerates into a direct bisection of the original PLM features. This prevents the "semantic collapse" often observed when randomly initialized complex layers are applied, gracefully preserving the fundamental aspect-level knowledge while enabling the network to learn smooth phase-shifting for polarity disentanglement.

\subsection{Joint Optimization Objective}
Our joint objective aligns semantic similarity, repels conflicting polarities, and prevents false-negative collisions.

\paragraph{Cosine Stabilization Objective:} 
While angle optimization focuses on phase disentanglement, we employ a standard InfoNCE-style cosine loss to stabilize the global aspect-level semantic clustering. It maximizes the similarity between the query $z$ and its positive target $w$ while repelling in-batch negatives:
\begin{equation}
\mathcal{L}_{cos} = - \log \frac{\exp(\cos(z, w) \cdot \tau_{cos})}{\sum_{n \in \mathcal{N}} \exp(\cos(z, n) \cdot \tau_{cos})}
\label{eq:cos_loss}
\end{equation}
where $\tau_{cos}$ acts as an inverse temperature scaling factor to control the penalty strength on hard negatives.

\paragraph{Masked In-Batch Negative Objective:} Standard contrastive learning erroneously repels same-class samples in a mini-batch ("False Negative Collisions"). We apply an anti-collision mask matrix $M$ to in-batch pairs based on polarity labels:

\begin{equation}
\begin{split}
\mathcal{L}_{\text{ibn}} \! = \! & -\sum_{i=1}^N \Big\{ \log \exp(s_i^+ \tau_{\text{ibn}}) \\
& - \log \big( \! \textstyle \sum_{j=1}^N \! \exp(s_{ij} \tau_{\text{ibn}}) (1 \! - \! M_{ij}) \big) \Big\}
\end{split}
\label{eq:ibn_loss}
\end{equation}

\paragraph{Angle-Optimized Objective:} 
Standard cosine similarity inherently suffers from \textit{gradient vanishing} in saturation zones ($|\sin\theta| \to 0$ as $\theta \to 0^\circ$), which prevents PLMs from repelling conflicting polarities within dense same-aspect clusters (Figure \ref{fig:gradient_vertical}a). To mitigate this gradient vanishing issue, we explicitly optimize the complex angular difference. Given query $z = a + bi$ and target $w = c + di$, the complex quotient reflects their phase divergence:
\begin{equation}
\frac{z}{w} \! = \! \frac{(a+bi)(c-di)}{(c+di)(c-di)} \! = \! \frac{ac+bd}{c^2+d^2} \! + \! i\frac{bc-ad}{c^2+d^2}
\label{eq:complex_div}
\end{equation}
To isolate sentiment disentanglement, we perform amplitude normalization to focus purely on the phase difference $\Delta\theta_{zw}$. Unlike the cosine objective, our angle-driven loss maintains a \textbf{constant gradient driving force} ($|\partial_\theta| = \text{Constant}$), ensuring robust separation of hard negatives regardless of their initial proximity (Figure \ref{fig:gradient_vertical}b):
\begin{equation}
\begin{split}
\mathcal{L}_{\text{angle}} \! = \! & \log \! \bigg( 1 + \! \sum_{n \in \mathcal{N}} \exp \Big( \Delta\theta_{zn} \cdot \tau_{\text{angle}} \\
& - \Delta\theta_{zw} \cdot \tau_{\text{angle}} \Big) \bigg)
\end{split}
\label{eq:angle_loss}
\end{equation}

\paragraph{Amplitude Consistency Objective:} 
To further enforce the structural consistency within the same aspect category, we introduce an \textbf{Amplitude Penalty ($\mathcal{L}_{amp}$)}. This objective constrains the complex magnitudes of the query $z$ and its target $w$ (whether positive or negative in polarity, as they share the same aspect) to align, ensuring that representations sharing the same objective aspect reside on a consistent hypersphere in the complex plane, thereby filtering out subjective intensity noise:
\begin{equation}
\mathcal{L}_{amp} = \text{MSE}(|z|, |w|)
\label{eq:amp_loss}
\end{equation}

\paragraph{Final Combined Objective:} While angle optimization dominates polarity separation, cosine optimization stabilizes global aspect clustering. Thus, our model is trained via a linear combination of the four losses:

\begin{equation}
\begin{split}
\mathcal{L}_{total} = & w_{ibn}\mathcal{L}_{ibn} + w_{angle}\mathcal{L}_{angle} \\
& + w_{cos}\mathcal{L}_{cos} + w_{amp}\mathcal{L}_{amp}
\end{split}
\label{eq:total_loss}
\end{equation}

In our optimal configuration \textbf{for high-density datasets}, weights are symmetrically balanced ($w_{\text{ibn}} = w_{\text{angle}} = w_{\text{cos}} = w_{\text{amp}} = 1.0$) to balance aggressive polarity disentanglement with fundamental semantic stability \textbf{(see Section \ref{sec:cross_lingual} for adaptations in sparse feature manifolds)}. This symmetric weighting implicitly acts as a geometric regularizer: while the angle loss aggressively separates polarities in the phase domain, the cosine and IBN losses anchor the vectors to prevent severe spatial fragmentation, maintaining an equilibrium between local intra-aspect cohesion and global inter-polarity repulsion.

% ==========================================
% 5. Experiment
% ==========================================

\section{Experiments}
\label{sec:experiments}
\subsection{Experimental Setup}
We evaluate our framework on the ASAP restaurant review dataset \cite{2019RoBERTa}, which contains 18 aspects with ternary polarities. Following the data disentanglement protocol, we extract aspect-specific triplets for training. The model is implemented using \texttt{chinese-roberta-wwm-ext} as the backbone. Key hyperparameters include: 5 training epochs, a learning rate of 2e-5 with a 500-step linear warmup, an effective batch size of 128. All experiments are conducted using FP16 mixed precision to optimize computational efficiency.

\subsection{Baselines}
We compare our proposed framework with several strong representation learning and fine-grained classification baselines, which can be logically categorized into three groups:

\paragraph{1. Foundation PLM \& Probing:}
\begin{itemize}[nosep, leftmargin=*]
    \item \textbf{RoBERTa (Zero-shot)}\cite{2019RoBERTa}: The standard pre-trained Chinese RoBERTa without fine-tuning, serving as the absolute lower bound.
    \item \textbf{RoBERTa-pair (Linear Probe)}\cite{sun2019utilizing}: A fundamental adaptation baseline that extracts \texttt{[CLS]} embeddings using aspect-context paired prompts. It serves as the baseline for frozen feature extraction.
\end{itemize}

\paragraph{2. General Contrastive Representation:}
\begin{itemize}[nosep, leftmargin=*]
    \item \textbf{SimCSE}\cite{gao2021simcse}: A state-of-the-art contrastive learning framework using dropout as minimal data augmentation, representing robust general semantic alignment.
\end{itemize}

\paragraph{3. Task-Specific ABSA Architectures:}
\begin{itemize}[nosep, leftmargin=*]
    \item \textbf{LCFS-RoBERTa}\cite{phan2020modelling}: A classical local context-focused model that isolates aspect-specific semantics by concatenating global representation with local context pooling.
    \item \textbf{DualGCN-RoBERTa}\cite{li-etal-2021-dual-graph}: A state-of-the-art syntax-aware model utilizing dual graph convolutional networks to capture complex syntactic dependencies.
    \item \textbf{Standard AnglE (Format C)}\cite{li2024angleoptimizedtextembeddings}: The original Angle-optimized text embeddings using direct hard chunking and standard triplets without anti-collision masks.
\end{itemize}

\subsection{Main Results}

% ==========================================
% 表格 1：主实验结果大表
% ==========================================

\begin{table*}[t]
\centering
\caption{Performance comparison on the ASAP dataset. Best results are \textbf{bolded}; second-best are \dotuline{underlined}.}
\label{tab:main_results}
\resizebox{\textwidth}{!}{
\begin{tabular}{l|cc|cc|cc|cc|cc|cc|cc}
\toprule
\multirow{2}{*}{\textbf{Aspect}} & \multicolumn{2}{c|}{\textbf{Zero-shot}} & \multicolumn{2}{c|}{\textbf{RoB-pair}} & \multicolumn{2}{c|}{\textbf{LCFS}} & \multicolumn{2}{c|}{\textbf{DualGCN}} & \multicolumn{2}{c|}{\textbf{SimCSE}} & \multicolumn{2}{c|}{\textbf{AnglE (Form C)}} & \multicolumn{2}{c}{\textbf{Ours (AnglE+ZRCP)}} \\
\cmidrule{2-15}
& F1 & Acc & F1 & Acc & F1 & Acc & F1 & Acc & F1 & Acc & F1 & Acc & F1 & Acc \\
\midrule
Transportation & 0.5359 & 0.6950 & 0.6559 & 0.8571 & 0.6834 & 0.8861 & 0.7950 & 0.9323 & \textbf{0.9039} & \dotuline{0.9710} & 0.8622 & 0.9592 & \dotuline{0.9022} & \textbf{0.9721} \\
Downtown & 0.4706 & 0.7217 & 0.5700 & 0.9245 & 0.6084 & 0.9363 & 0.7454 & 0.9658 & \dotuline{0.7961} & 0.9682 & \textbf{0.8434} & \textbf{0.9776} & 0.7858 & \dotuline{0.9717} \\
Easy to find & 0.5998 & 0.6601 & 0.7150 & 0.7696 & 0.7534 & 0.8076 & 0.9004 & 0.9297 & \dotuline{0.9191} & \dotuline{0.9424} & 0.9149 & 0.9401 & \textbf{0.9265} & \textbf{0.9482} \\
Queue & 0.6830 & 0.6851 & 0.7502 & 0.7511 & 0.7649 & 0.7660 & 0.7802 & 0.7809 & 0.8401 & 0.8404 & \dotuline{0.8573} & \dotuline{0.8574} & \textbf{0.8743} & \textbf{0.8745} \\
Hospitality & 0.7932 & 0.8526 & 0.8678 & 0.9039 & 0.8793 & 0.9138 & 0.9281 & 0.9497 & \dotuline{0.9505} & \dotuline{0.9660} & 0.9479 & 0.9642 & \textbf{0.9582} & \textbf{0.9714} \\
Parking & 0.6495 & 0.6809 & 0.6570 & 0.7121 & 0.6720 & 0.7354 & 0.8319 & 0.8599 & \dotuline{0.9072} & \dotuline{0.9261} & 0.8965 & 0.9183 & \textbf{0.9169} & \textbf{0.9339} \\
Timely & 0.6580 & 0.6733 & 0.8146 & 0.8222 & 0.8210 & 0.8283 & 0.8816 & 0.8860 & 0.9161 & 0.9195 & \textbf{0.9414} & \textbf{0.9438} & \dotuline{0.9329} & \dotuline{0.9362} \\
Price Level & 0.6645 & 0.6662 & 0.7628 & 0.7634 & 0.7824 & 0.7826 & 0.9015 & 0.9021 & 0.9312 & 0.9318 & \dotuline{0.9335} & \dotuline{0.9340} & \textbf{0.9342} & \textbf{0.9347} \\
Cost-effective & 0.6600 & 0.7696 & 0.7522 & 0.8514 & 0.7689 & 0.8684 & 0.8732 & 0.9352 & 0.9035 & 0.9512 & \textbf{0.9223} & \textbf{0.9628} & \dotuline{0.9212} & \dotuline{0.9607} \\
Discount & 0.5751 & 0.6994 & 0.6975 & 0.8264 & 0.7248 & 0.8604 & 0.7146 & 0.8453 & 0.7347 & 0.8654 & \dotuline{0.7744} & \textbf{0.9006} & \textbf{0.7835} & \dotuline{0.8956} \\
Decoration & 0.5841 & 0.7759 & 0.6891 & 0.8808 & 0.7334 & 0.9113 & 0.7814 & 0.9343 & 0.8057 & 0.9412 & \dotuline{0.8281} & \dotuline{0.9539} & \textbf{0.8627} & \textbf{0.9637} \\
Noise & 0.6127 & 0.7136 & 0.7539 & 0.8593 & 0.7714 & 0.8751 & 0.8840 & 0.9417 & 0.9079 & 0.9550 & \textbf{0.9306} & \textbf{0.9667} & \dotuline{0.9272} & \dotuline{0.9642} \\
Space & 0.6557 & 0.7050 & 0.7695 & 0.8245 & 0.7710 & 0.8306 & 0.8805 & 0.9160 & \dotuline{0.8809} & \dotuline{0.9168} & 0.8774 & 0.9160 & \textbf{0.9108} & \textbf{0.9380} \\
Sanitary & 0.7133 & 0.8106 & 0.7862 & 0.8641 & 0.8072 & 0.8838 & \dotuline{0.8776} & \dotuline{0.9289} & 0.8654 & 0.9197 & 0.8744 & 0.9296 & \textbf{0.9002} & \textbf{0.9423} \\
Portion & 0.6490 & 0.6916 & 0.7347 & 0.7710 & 0.7690 & 0.8052 & 0.8531 & 0.8788 & \dotuline{0.8864} & \dotuline{0.9072} & 0.8776 & 0.9020 & \textbf{0.9013} & \textbf{0.9200} \\
Taste & 0.7500 & 0.9001 & 0.8538 & 0.9522 & 0.8896 & 0.9667 & 0.8809 & 0.9624 & \dotuline{0.9048} & \dotuline{0.9710} & 0.8973 & 0.9685 & \textbf{0.9247} & \textbf{0.9776} \\
Appearance & 0.6823 & 0.7996 & 0.7224 & 0.8363 & 0.7390 & 0.8561 & 0.7721 & 0.8692 & 0.8022 & 0.8899 & \textbf{0.8366} & \textbf{0.9182} & \dotuline{0.8105} & \dotuline{0.8975} \\
Recommend & 0.6094 & 0.7194 & 0.6913 & 0.8189 & 0.7187 & 0.8444 & 0.7951 & 0.8941 & \textbf{0.8911} & \textbf{0.9515} & 0.8825 & 0.9490 & \dotuline{0.8887} & \dotuline{0.9503} \\
\midrule
\textbf{Macro-Avg} & 0.6414 & 0.7344 & 0.7358 & 0.8327 & 0.7588 & 0.8532 & 0.8376 & 0.9062 & \dotuline{0.8748} & \dotuline{0.9297} & 0.8823 & 0.9340 & \textbf{0.8923} & \textbf{0.9418} \\
\bottomrule
\end{tabular}
}
\end{table*}

Table \ref{tab:main_results} presents the performance of all models across 18 specific aspects. Our proposed framework consistently outperforms all competitive baselines, achieving a best Macro-F1 score of \textbf{0.8923} and an overall Accuracy of \textbf{0.9418}.

\paragraph{Comparison with Real-Valued Baselines:}
As observed, while foundational models like RoB-pair (0.7358) and task-specific architectures like LCFS (0.7588) and DualGCN (0.8376) yield significant improvements over Zero-shot RoBERTa, they still fall short of our framework. This performance gap suggests that mapping fine-grained aspect-sentiment pairs into a standard real-valued space inherently limits the model's ability to draw crisp decision boundaries. Even with manual architectural priors (e.g., syntax graphs in DualGCN), these models suffer from representation entanglement, especially when processing complex subjective expressions.

\paragraph{Comparison with Contrastive Baselines:}
Our model outperforms general contrastive baselines, including SimCSE (0.8748) and the standard AnglE (0.8823). Notably, compared to the unmasked AnglE baseline, our framework achieves substantial gains in challenging and long-tail categories. For instance, in the \textit{Downtown} aspect, our model improves the F1 score to \textbf{0.7858}, demonstrating the effectiveness of our joint optimization strategy in expanding discriminative margins. These results validate that our approach provides a more robust and aligned representation space for ABSA. A detailed analysis of the individual module contributions and geometric properties is provided in Section \ref{sec:amp_analysis}.

\subsection{Ablation Study}
\label{sec:ablation}

To verify the effectiveness of our proposed modules and data strategies, we conducted an extensive ablation study by systematically removing key components from our full framework. The results are summarized in Table \ref{tab:ablation}.

% ==========================================
% 表格 2：消融实验表 (包含架构与数据策略)
% ==========================================
\begin{table}[htbp]
\centering
% 自适应宽度：缩放到适合单栏的大小
\resizebox{\linewidth}{!}{%
\begin{tabular}{lcc}
\toprule
\textbf{Model Variations} & \textbf{Macro-F1} & \textbf{Acc} \\
\midrule
\textbf{Ours (Full Framework)} & \textbf{0.8923} & \textbf{0.9418} \\
\midrule
\multicolumn{3}{l}{\textit{Ablation on Network \& Objective}} \\
\quad $w/o$ ZRCP (Hard Chunking) & 0.8739 & 0.9319 \\
\quad $w/o$ Anti-collision Mask & 0.8809 & 0.9336 \\
\quad $w/o$ Angle Loss ($w_{angle}=0$) & 0.8843 & 0.9389 \\
\midrule
\multicolumn{3}{l}{\textit{Ablation on Data Strategy}} \\
\quad $w/o$ Dynamic Window (Full Text) & 0.8679 & 0.9283 \\
\quad $w/o$ Hybrid Triplets (Standard Pairs) & 0.7735 & 0.8607 \\
\bottomrule
\end{tabular}
}
\caption{Ablation study of architectural components and data strategies.}
\label{tab:ablation}
\end{table}

\begin{figure*}[htbp]
    \centering
    % 请将文件名替换为你保存的图片名
    \includegraphics[width=0.9\linewidth]{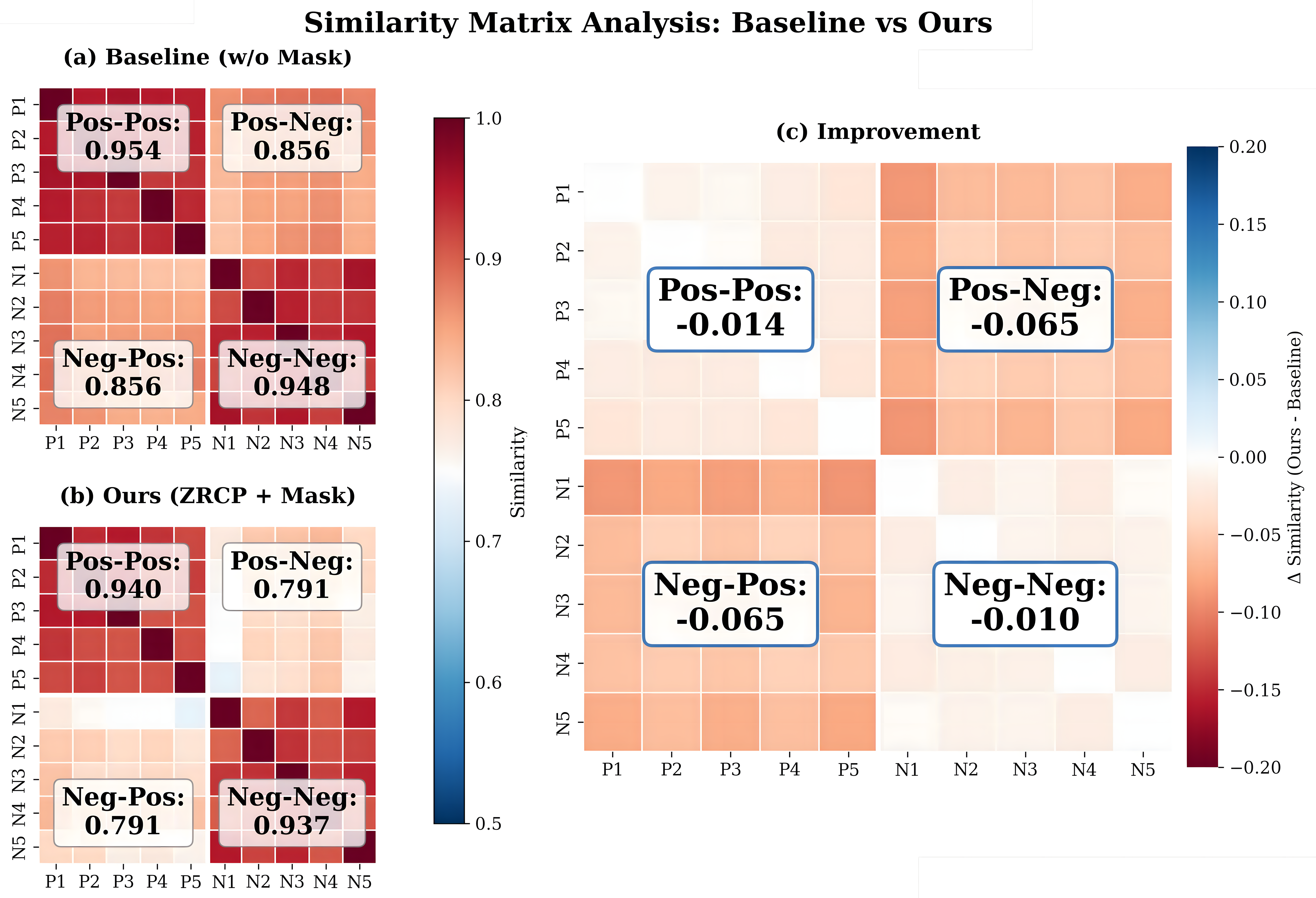}
    \caption{Similarity matrix analysis. (a) The unmasked baseline suffers from false negative collisions. (b) Our method delineates clear block-diagonal structures. (c) Our model expands the inter-polarity discriminative margin ($-6.5\%$) while preserving intra-polarity cohesion.}\label{fig:similarity_analysis}
\end{figure*}

\paragraph{Ablation on Network \& Objective:}
We first analyze the contribution of each architectural component. 
\begin{itemize}[nosep, leftmargin=*]
    \item \textbf{$w/o$ ZRCP}: Replacing our zero-initialized residual projection with direct tensor chunking leads to a significant drop in Macro-F1 (from 0.8923 to 0.8739). This confirms that hard bisection disrupts the semantic continuity of pre-trained embeddings, whereas ZRCP preserves the fundamental representation while enabling smooth complex mapping.
    \item \textbf{$w/o$ Anti-collision Mask}: Reverting to standard triplet objectives causes the Macro-F1 to decline to 0.8809. This drop is primarily due to false-negative collisions in high-frequency aspects, where same-polarity samples are erroneously repelled.
    \item \textbf{$w/o$ Angle Loss}: When $w_{angle}=0$, the model relies solely on cosine similarity, resulting in a Macro-F1 of 0.8843. This validates that cosine similarity alone cannot overcome the gradient vanishing issue in high-density pre-trained clusters, whereas our angle objective provides the necessary driving force for polarity separation.
\end{itemize}

\paragraph{Ablation on Data Strategy:}
As shown in Table \ref{tab:ablation}, our data-level disentanglement is equally vital. 
\begin{itemize}[nosep, leftmargin=*]
    \item \textbf{$w/o$ Dynamic Window}: Forcing the model to process full, noisy reviews reduces the Macro-F1 to 0.8679. This proves that bounding the context is essential to prevent extraneous sentiments from polluting aspect-specific representations.
    \item \textbf{$w/o$ Hybrid Triplets}: Replacing our hybrid mining with standard random pairs results in a significant performance degradation, with Macro-F1 plummeting to 0.7735. This underscores that our phase-driven architecture requires high-density contrastive signals to fully activate its disentanglement capabilities.
\end{itemize}

\subsection{Cross-lingual Generalization}
\label{sec:cross_lingual}

To verify the language-agnostic robustness and the disentanglement capability of our geometric framework, we evaluate the model on two challenging English datasets: SemEval-2016 (Task 5) \cite{pontiki-etal-2016-semeval} and MAMS-ACSA \cite{jiang-etal-2019-challenge}. Compared to Chinese reviews, SemEval-2016 exhibits severe class imbalance. Furthermore, MAMS-ACSA provides a uniquely rigorous testbed for representation entanglement, as it guarantees that every sentence contains at least two aspects with different sentiment polarities, heavily penalizing models that rely on global sentence-level sentiment representations.

We adapt our architecture using the English \texttt{roberta-base} backbone. As shown in Table \ref{tab:combined_results}, our ZRCP framework establishes state-of-the-art performance across both benchmarks, achieving an outstanding Macro-F1 of 0.9044 on SemEval-2016 and 0.6671 on MAMS-ACSA. 

We observe that the English feature space is notably more compact than its Chinese counterpart. In this sparse regime, we identify a critical Uniformity Trade-off: while the anti-collision mask is essential for dense datasets, the "false-negative collisions" in standard contrastive learning paradoxically act as a vital \textit{uniformity regularizer} for small-scale manifolds, preventing representations from collapsing into singular points. Consequently, for English adaptation, we treat $\mathcal{L}_{angle}$ and $\mathcal{L}_{amp}$ as soft regularizers ($w=0.02$) and disable the hard mask penalty to maintain sufficient repulsive tension. This configuration effectively prevents manifold tearing while rescuing catastrophic spatial collapse on long-tail aspects, such as improving the \textit{Location} F1 from 0.3636 to 1.0000 on SemEval-2016 (noting that the absolute score peak is partially attributed to the extremely small sample size of this specific category in the test set). Detailed per-aspect metrics for both datasets are documented in Appendix \ref{sec:appendix_cross_lingual_details}.

\begin{table}[htbp]
\centering
\caption{Overall performance comparison on SemEval-2016 and MAMS-ACSA datasets. Best results are \textbf{bolded}; second-best are \dotuline{underlined}.}
\label{tab:combined_results}
\resizebox{\linewidth}{!}{
\begin{tabular}{l|cc|cc}
\toprule
\multirow{2}{*}{\textbf{Model}} & \multicolumn{2}{c|}{\textbf{SemEval-2016}} & \multicolumn{2}{c}{\textbf{MAMS-ACSA}} \\
\cmidrule{2-5}
& \textbf{Macro-F1} & \textbf{Acc} & \textbf{Macro-F1} & \textbf{Acc} \\
\midrule
RoB-pair & 0.8180 & 0.8869 & 0.6383 & 0.7193 \\
DualGCN & 0.8289 & 0.9135 & 62.643 & 66.548 \\
SimCSE & 0.8457 & 0.8962 & 0.6038 & 0.7217 \\
AnglE (Format C) & 0.8632 & 0.9134 & -- & -- \\
AnglE (Format A) & \dotuline{0.8784} & \dotuline{0.9078} & \dotuline{0.6532} & \dotuline{0.7526} \\
\midrule
\textbf{Ours (ZRCP+Mask)} & \textbf{0.9044} & \textbf{0.9347} & \textbf{0.6671} & \textbf{0.7656} \\
\bottomrule
\end{tabular}
}
\end{table}

\begin{figure*}[t]
    \centering
    \includegraphics[width=0.85\linewidth]{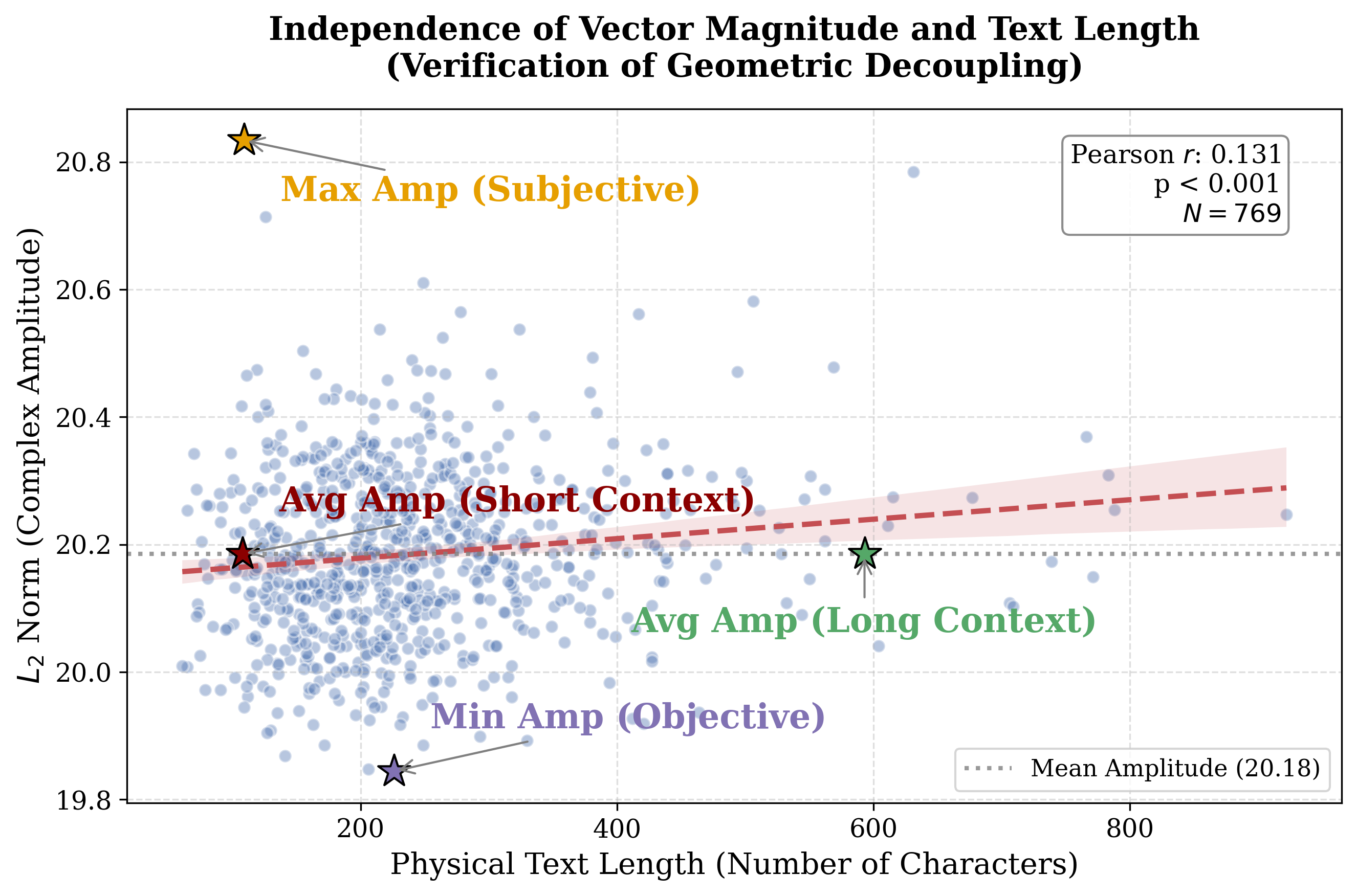}
    \caption{Statistical verification of length-amplitude decoupling ($r=0.130, p<0.001$), demonstrating robustness against physical text length biases.}
    \label{fig:amp_geo}
\end{figure*}

\subsection{Representation Space Analysis}
\label{sec:space_analysis}

While global metrics like Macro-F1 demonstrate quantitative improvements (Section 5.3), the true geometric advantage of our framework is best visualized through batch-level similarity structures. To gain deeper insights into how our joint objective refines the discriminative margin, we visualize both the pairwise cosine similarity heatmaps and the aggregated mean similarities for an aspect-matched batch (e.g., 5 positive and 5 negative sentences sharing the ``Taste'' aspect, detailed in Appendix \ref{sec:appendix_similarity_samples}).

As illustrated in Figure \ref{fig:similarity_analysis}(a), in the baseline model trained without anti-collision masks (Format C), the semantic space is chaotic. False negative collisions erroneously repel sentences with identical sentiments, destroying intra-class cohesion and blurring the decision boundary. 

In stark contrast, our ZRCP+Mask framework (Figure \ref{fig:similarity_analysis}(b)) presents delineated block-diagonal structures. Quantitative analysis of the aggregated similarities (Figure \ref{fig:similarity_analysis}(c)) confirms that our method slightly reduces the 'Pos-Pos' and 'Neg-Neg' intra-polarity cohesion from 0.954 to 0.940 (-1.4\%) and from 0.948 to 0.937 (-1.0\%) respectively. In fine-grained ABSA, this slight reduction acts as a vital geometric trade-off. Sentences sharing the same aspect naturally exhibit high baseline similarities; forcing them closer would lead to spatial collapse. Instead, sacrificing a fraction of this cohesion enables the network to significantly suppress the false similarities between opposing polarities. 

This trade-off aligns with the fundamental distinction between fine-grained sentiment analysis and general semantic similarity tasks: what matters most is not how similar identical polarities are, but how distinct opposing polarities become.

Specifically, the mean 'Pos-Neg' inter-polarity similarity is dramatically reduced from 0.856 to 0.791, achieving a notable 6.5\% distinction. This empirical evidence confirms that our framework expands the discriminative margin to a healthy level, effectively mitigating the information hallucination observed in the baseline, where opposing sentiments are erroneously rendered almost as similar as identical sentiments.

\subsection{Deep Analysis: Structural Consolidation via Amplitude Penalty}
\label{sec:amp_analysis}

In Section 4.2, we introduced the Amplitude Penalty ($\mathcal{L}_{amp}$) to stabilize aspect representations. To understand its geometric impact, we compare our full framework against a version where $w_{amp}=0$. As shown in Table \ref{tab:amp_analysis}, adding $\mathcal{L}_{amp}$ optimizes the model's structure: the overall Accuracy remains remarkably stable (94.18\% vs. 94.19\%), while the Macro-F1 achieves a consistent gain, rising from 0.8901 to 0.8923.

\begin{table}[htbp]
\centering
\small
\resizebox{\linewidth}{!}{
\begin{tabular}{lccc}
\toprule
\textbf{Aspect Category} & \textbf{w/o AmpLoss} & \textbf{with AmpLoss} & \textbf{$\Delta$} \\
\midrule
\multicolumn{4}{c}{\textit{Objective / Fact-based Aspects}} \\
Transportation & 0.8916 & 0.9022 & +1.06\% \\
Downtown & 0.7489 & 0.7858 & +3.69\% \\
\midrule
\multicolumn{4}{c}{\textit{Subjective / Descriptive Aspects}} \\
Decoration & 0.8556 & 0.8627 & +0.71\% \\
Taste & 0.9265 & 0.9247 & -0.18\% \\
\midrule
\textbf{Macro-Average F1} & \textbf{0.8901} & \textbf{0.8923} & \textbf{+0.22\%} \\
\textbf{Accuracy} & \textbf{0.9419} & \textbf{0.9418} & \textbf{-0.01\%} \\
\bottomrule
\end{tabular}
}
\caption{Impact of Amplitude Penalty (\texttt{AmpLoss}) on divergent aspects.}
\label{tab:amp_analysis}
\end{table}

This phenomenon suggests that $\mathcal{L}_{amp}$ acts as a powerful \textbf{structural regularizer}. By constraining vectors onto a unified hypersphere, the model filters out extraneous ``intensity noise'' caused by subjective lexical richness, forcing the optimization to focus exclusively on phase-driven polarity. This is particularly evident in long-tail aspects like \textit{Downtown}, where F1 drastically improves by 3.69\%. This trade-off demonstrates that while maintaining global accuracy, the structural consolidation significantly enhances the model's robustness on class-imbalanced dimensions, preventing over-fitting to intensity-driven surface patterns.

% ==========================================
% 6. Conclusion
% ==========================================

\section{Conclusion}
We presented a novel complex-valued framework for ABSA, utilizing Zero-Initialized Residual Complex Projection (ZRCP) and an Anti-collision Masked Angle Loss to disentangle aspect and sentiment. Our geometric approach effectively overcomes the gradient limitations of cosine similarity and mitigates representation entanglement. Experimental results on the ASAP dataset demonstrate the superiority of our method, particularly in expanding inter-polarity margins while preserving semantic stability. Future work will explore the extension of this geometric paradigm to multimodal sentiment analysis.

% ==========================================
% 7. Limitations
% ==========================================
\section*{Limitations}

\label{sec:limitations}
Despite the significant performance gains, several limitations remain. First, our context-aware aspect extraction algorithm partially relies on external keyword lists and lexicons to locate aspect anchors. In domains where such linguistic resources are scarce, the precision of anchor localization and subsequent context window extraction may decrease. Second, the joint optimization objective involves multiple hyperparameters ($w_{ibn}$, $w_{angle}$, $w_{cos}$) and temperature parameters ($\tau$). Finding the optimal balance between these objectives can be computationally expensive. Third, the ZRCP module introduces additional linear projection layers and complex-valued operations, which leads to a slight increase in computational overhead and inference latency compared to standard real-valued models. Finally, our dynamic windowing strategy currently employs a localized context of 2 to 4 sentences; for exceptionally long reviews with complex logical spans, this local approach might overlook long-range semantic dependencies.

\bibliography{custom}

\clearpage
\appendix

% ==========================================
% Table: SemEval-2016 Detailed Results (Appendix)
% ==========================================

\begin{table*}[htbp]
\centering
\caption{Detailed per-aspect performance on the English SemEval-2016 dataset. Best results are \textbf{bolded}; second-best are \dotuline{underlined}.}
\label{tab:semeval_detailed_appendix}
\resizebox{\textwidth}{!}{
\begin{tabular}{l|cc|cc|cc|cc|cc|cc}
\toprule
\multirow{2}{*}{\textbf{Aspect}} & \multicolumn{2}{c|}{\textbf{RoB-pair}} & \multicolumn{2}{c|}{\textbf{DualGCN}} & \multicolumn{2}{c|}{\textbf{SimCSE}} & \multicolumn{2}{c|}{\textbf{AnglE (Format C)}} & \multicolumn{2}{c|}{\textbf{AnglE (Format A)}} & \multicolumn{2}{c}{\textbf{Ours (ZRCP+Mask)}} \\
\cmidrule{2-13}
& F1 & Acc & F1 & Acc & F1 & Acc & F1 & Acc & F1 & Acc & F1 & Acc \\
\midrule
Restaurant General & 0.9549 & 0.9701 & 0.8923 & 0.9220 & 0.9780 & 0.9851 & 0.9780 & 0.9851 & \textbf{1.0000} & \textbf{1.0000} & 0.9780 & 0.9851 \\
Service & \textbf{1.0000} & \textbf{1.0000} & 0.9122 & 0.9122 & 0.9787 & 0.9825 & 0.9581 & 0.9649 & 0.9787 & 0.9825 & 0.9566 & 0.9649 \\
Food Quality & 0.9424 & 0.9737 & 0.8493 & 0.9433 & 0.8848 & 0.9474 & \textbf{0.9699} & \textbf{0.9868} & \textbf{0.9699} & \textbf{0.9868} & 0.9171 & 0.9605 \\
Food Style/Options & 0.7013 & 0.7826 & 0.8627 & 0.8913 & 0.7319 & 0.7826 & \textbf{0.7745} & \textbf{0.8261} & \textbf{0.7745} & \textbf{0.8261} & 0.7319 & 0.7826 \\
Food Prices & 0.8611 & 0.8667 & 0.8400 & 0.8500 & 0.9321 & 0.9333 & 0.9321 & 0.9333 & \textbf{1.0000} & \textbf{1.0000} & 0.9321 & 0.9333 \\
Drinks Quality & 1.0000 & 1.0000 & 0.7250 & 0.9091 & \textbf{1.0000} & \textbf{1.0000} & \textbf{1.0000} & \textbf{1.0000} & \textbf{1.0000} & \textbf{1.0000} & \textbf{1.0000} & \textbf{1.0000} \\
Drinks Prices & 1.0000 & 1.0000 & 1.0000 & 1.0000 & \textbf{1.0000} & \textbf{1.0000} & \textbf{1.0000} & \textbf{1.0000} & \textbf{1.0000} & \textbf{1.0000} & \textbf{1.0000} & \textbf{1.0000} \\
Ambience & 0.4603 & 0.8529 & 0.7009 & 0.9206 & 0.7344 & 0.9412 & 0.8256 & 0.9706 & \textbf{1.0000} & \textbf{1.0000} & 0.7344 & 0.9412 \\
Location & 0.3636 & 0.5714 & 0.4762 & 0.9091 & 0.3636 & 0.5714 & 0.4167 & 0.7143 & 0.3636 & 0.5714 & \textbf{1.0000} & \textbf{1.0000} \\
Restaurant Prices & 0.9282 & 0.9286 & 0.8348 & 0.8421 & \dotuline{0.8542} & \dotuline{0.8571} & \dotuline{0.8542} & \dotuline{0.8571} & \dotuline{0.8542} & \dotuline{0.8571} & \dotuline{0.8542} & \dotuline{0.8571} \\
Restaurant Misc & 0.7857 & 0.8095 & 0.8535 & 0.8621 & \textbf{0.8444} & \textbf{0.8571} & 0.7857 & 0.8095 & 0.7215 & 0.7619 & \textbf{0.8444} & \textbf{0.8571} \\
\midrule
\textbf{Macro-Avg} & 0.8180 & 0.8869 & 0.8289 & 0.9135 & 0.8457 & 0.8962 & 0.8632 & 0.9134 & \dotuline{0.8784} & \dotuline{0.9078} & \textbf{0.9044} & \textbf{0.9347} \\
\bottomrule
\end{tabular}
}
\end{table*}

\begin{table*}[htbp]
\centering
\caption{Detailed per-aspect performance on the MAMS-ACSA dataset. Best results are \textbf{bolded} and second-best are \dotuline{underlined}.}
\label{tab:mams_detailed_appendix}
\resizebox{\textwidth}{!}{
\begin{tabular}{l|cc|cc|cc|cc|cc}
\toprule
\multirow{2}{*}{\textbf{Aspect}} & \multicolumn{2}{c|}{\textbf{RoB-pair}} & \multicolumn{2}{c|}{\textbf{DualGCN}} & \multicolumn{2}{c|}{\textbf{SimCSE}} & \multicolumn{2}{c|}{\textbf{AnglE}} & \multicolumn{2}{c}{\textbf{Ours (AnglE+ZRCP)}} \\
\cmidrule{2-11}
& F1 & Acc & F1 & Acc & F1 & Acc & F1 & Acc & F1 & Acc \\
\midrule
Food & \textbf{0.7298} & \dotuline{0.7526} & 0.5237 & 0.5464 & 0.6008 & 0.6564 & 0.6920 & 0.7354 & \dotuline{0.7224} & \textbf{0.7698} \\
Service & \dotuline{0.7702} & 0.8077 & 0.5771 & 0.7308 & 0.6737 & 0.7308 & 0.7676 & \dotuline{0.8205} & \textbf{0.8384} & \textbf{0.8590} \\
Staff & 0.6108 & 0.7870 & 0.5925 & \textbf{0.8284} & \dotuline{0.6166} & \dotuline{0.8047} & 0.5995 & \dotuline{0.8047} & \textbf{0.6199} & \dotuline{0.8047} \\
Price & \dotuline{0.7540} & \dotuline{0.7368} & 0.6735 & 0.6579 & 0.6647 & 0.6579 & \textbf{0.7868} & \textbf{0.7895} & 0.7210 & 0.7105 \\
Ambience & 0.5302 & 0.7188 & 0.5494 & \dotuline{0.8125} & \textbf{0.5703} & \textbf{0.8438} & 0.5297 & 0.7188 & \dotuline{0.5523} & 0.7500 \\
Menu & 0.4849 & 0.6579 & \textbf{0.7793} & \textbf{0.9211} & 0.5261 & 0.7763 & 0.5542 & 0.7763 & \dotuline{0.6483} & \dotuline{0.8553} \\
Place & 0.6474 & 0.6543 & 0.4318 & 0.4444 & 0.6240 & \dotuline{0.6790} & \textbf{0.7012} & \textbf{0.7284} & \dotuline{0.6723} & \textbf{0.7284} \\
Miscellaneous & \dotuline{0.5788} & \dotuline{0.6397} & 0.3935 & 0.3824 & 0.5544 & 0.6250 & \textbf{0.5942} & \textbf{0.6471} & 0.5621 & \textbf{0.6471} \\
\midrule
\textbf{Macro-Avg} & 0.6383 & 0.7193 & 0.5651 & 0.6655 & 0.6038 & 0.7217 & \dotuline{0.6532} & \dotuline{0.7526} & \textbf{0.6671} & \textbf{0.7656} \\
\bottomrule
\end{tabular}
}
\end{table*}

\section{Detailed Results on Cross-lingual Datasets}
\label{sec:appendix_cross_lingual_details}

To support the cross-lingual generalization claims discussed in Section \ref{sec:cross_lingual}, we provide the comprehensive, per-aspect evaluation results on both the English SemEval-2016 (Task 5) dataset and the MAMS-ACSA dataset.

\paragraph{Analysis on SemEval-2016:}
As shown in Table \ref{tab:semeval_detailed_appendix}, the detailed metrics highlight the "seesaw effect" inherent in deep geometric representation learning. While strong contrastive baselines like SimCSE and AnglE achieve marginal advantages on high-frequency, relatively simple aspects (e.g., \textit{Food Quality}), they experience severe spatial collapse on sparse, long-tail aspects (e.g., \textit{Location} and \textit{Food Style/Options}). Our proposed ZRCP framework, equipped with the amplitude penalty, structurally consolidates the representation space. By anchoring vectors onto a unified hypersphere, our model successfully rescues these collapsed long-tail categories, elevating the F1 score of \textit{Location} from 0.3636 to 1.0000 (noting that the absolute score peak is partially attributed to the extremely small sample size of this specific category in the test set), and ultimately achieving a superior global Macro-F1 of 0.9044.

\paragraph{Analysis on MAMS-ACSA:}
To further demonstrate our framework's ability to decouple conflicting sentiments within the exact same context, we present the per-aspect results on the MAMS-ACSA dataset in Table \ref{tab:mams_detailed_appendix}. Because every sentence in MAMS contains multiple aspects with opposite polarities, real-valued task-specific models like DualGCN severely struggle with representation entanglement (yielding a Macro-F1 of only 0.5651). Conversely, our phase-driven disentanglement achieves a robust Macro-F1 of 0.6671. Notably, in highly subjective and frequently entangled categories such as \textit{Service} and \textit{Menu}, our method provides substantial performance margins. This empirically verifies that our framework effectively isolates conflicting polarities by strictly separating them via the phase domain while structurally constraining the objective aspects through the amplitude penalty.

\section{Representative Samples for Similarity Matrix Analysis}
\label{sec:appendix_similarity_samples}

To facilitate the qualitative understanding of the similarity matrix and discriminative margins discussed in Section \ref{sec:space_analysis}, we present the 10 representative samples (5 Positive, 5 Negative) used to generate Figure \ref{fig:similarity_analysis}. These samples evaluate the \textit{Taste} aspect and exhibit high semantic density, which typically induces false-negative collisions in standard contrastive baselines.

\paragraph{Positive Samples (Intra-polarity Cohesion)}
\begin{itemize}[nosep, leftmargin=*]
    \item \textbf{P1:} \begin{CJK*}{UTF8}{gbsn}“味道惊艳，回味无穷，绝对是我吃过最好吃的！”\end{CJK*} (Amazing taste with a long finish, absolutely the best I've had!)
    \item \textbf{P2:} \begin{CJK*}{UTF8}{gbsn}“完美！口感层次丰富，每一口都是享受，太赞了！”\end{CJK*} (Perfect! Rich layers of texture, every bite is a joy, so great!)
    \item \textbf{P3:} \begin{CJK*}{UTF8}{gbsn}“太绝了，从第一口到最后一口都超级满足，人间美味！”\end{CJK*} (Incredible, super satisfying from the first bite to the last, a delicacy!)
    \item \textbf{P4:} \begin{CJK*}{UTF8}{gbsn}“无可挑剔，色香味俱全，强烈推荐给所有人！”\end{CJK*} (Impeccable, looks, smells, and tastes great, highly recommended!)
    \item \textbf{P5:} \begin{CJK*}{UTF8}{gbsn}“满分好评，味道太赞了，下次一定还来，超预期！”\end{CJK*} (Full marks, taste is awesome, will definitely come back, exceeded expectations!)
\end{itemize}

\paragraph{Negative Samples (Inter-polarity Repulsion)}
\begin{itemize}[nosep, leftmargin=*]
    \item \textbf{N1:} \begin{CJK*}{UTF8}{gbsn}“太难吃了，像嚼蜡一样，完全没味道，恶心！”\end{CJK*} (Tastes terrible, like chewing wax, no flavor at all, disgusting!)
    \item \textbf{N2:} \begin{CJK*}{UTF8}{gbsn}“难以下咽，食材不新鲜，有怪味，吃一口就想吐。”\end{CJK*} (Hard to swallow, ingredients not fresh, has an odd smell, makes me want to vomit.)
    \item \textbf{N3:} \begin{CJK*}{UTF8}{gbsn}“太失望了，味道奇怪，根本吃不下去，浪费钱。”\end{CJK*} (So disappointed, strange taste, can't eat it at all, a waste of money.)
    \item \textbf{N4:} \begin{CJK*}{UTF8}{gbsn}“差评！又咸又腻，吃完反胃，绝对不会再来。”\end{CJK*} (Negative review! Too salty and greasy, feels nauseous after eating, will never return.)
    \item \textbf{N5:} \begin{CJK*}{UTF8}{gbsn}“口感极差，像是在吃塑料，太难吃了，后悔死了。”\end{CJK*} (Terrible texture, like eating plastic, so bad, I regret it so much.)
\end{itemize}

\section{Textual Case Studies of Amplitude Extremes}
\label{sec:appendix_case_study}

To provide a rigorous empirical foundation for the geometric properties and amplitude distribution discussed in Section \ref{sec:amp_analysis}, we present full texts of the extreme cases extracted from the test set (evaluating the \textit{Decoration} aspect). These examples empirically demonstrate how subjective intensity, rather than physical length, disrupts real-valued feature spaces.

\paragraph{Case 1: Minimum Amplitude ($|Z| \approx 20.04, \text{Length} = 71$)}
This instance is physically long but consists primarily of objective narrative listing decor items and location details. The lack of intense subjective evaluation results in a minimal amplitude in the unconstrained space.
\begin{itemize}[leftmargin=*]
    \item \textbf{Full Text:} \begin{CJK*}{UTF8}{gbsn}“\#烤匠\# 先说下环境，群光广场9楼找到店家，外面排队的人贼多贼多，还好我可以直接进来[调皮]，装修是酒吧格调，卡座和吧台，所以每桌人数不能太多”\end{CJK*}
    \item \textbf{Translation:} ``\#KaoJiang\# Let's talk about the environment first. Found the store on the 9th floor of Chicony Plaza. There were so many people waiting in line outside, but luckily I could come straight in. The decoration is a bar style, with booths and a bar counter, so the number of people at each table cannot be too many.''
\end{itemize}

\paragraph{Case 2: Maximum Amplitude ($|Z| \approx 20.84, \text{Length} = 19$)}
Although significantly shorter (only 19 characters), this text is saturated with high-intensity subjective descriptors ("very OK", "very comfortable"). These evaluations strongly stretch the complex magnitude in baseline models.
\begin{itemize}[leftmargin=*]
    \item \textbf{Full Text:} \begin{CJK*}{UTF8}{gbsn}“店里的环境者很OK，音乐和装修都很舒服”\end{CJK*}
    \item \textbf{Translation:} ``The environment in the store is very OK, and the music and decoration are very comfortable.''
\end{itemize}

\paragraph{Case 3: Amplitude vs. Length Decoupling Verification}
To further prove that vector magnitude operates independently of physical token counts, we highlight two additional instances sharing the exact same average amplitude ($|Z| \approx 20.19$) but exhibiting a $5.5\times$ difference in length. This empirical observation supports the low correlation ($r=0.130$) shown in Figure \ref{fig:amp_geo}.
\begin{itemize}[leftmargin=*]
    \item \textbf{Long Context (Length = 593):} \begin{CJK*}{UTF8}{gbsn}“我只想吐槽点评自带的定位和导航... 餐厅装修的很美式... 奶油白葡萄酒海虹—点评第一推荐菜...”\end{CJK*}
    \item \textbf{Short Context (Length = 108):} \begin{CJK*}{UTF8}{gbsn}“带有点乡土气息的华丽大酒店。装修过后，虽然门头还是不起眼，但是内部立马华丽了起来。”\end{CJK*}
\end{itemize}

\end{document}